# A Case Study of Chinese Sentiment Analysis of Social Media Reviews Based on LSTM

Lukai Wang[1], Lei Wang[2],[*]

[1]Department of Plasma Physics and Fusion Engineering and CAS Key Laboratory of Geospace Environment, University of Science and Technology of China, Hefei, Anhui 230026, People's Republic of China
[2]School of Foreign Languages, Peking University, Beijing 100871, China
[*]wangleics@pku.edu.cn

## Abstract

Network public opinion analysis is obtained by a combination of natural language processing (NLP) and public opinion supervision, and is crucial for monitoring public mood and trends. Therefore, network public opinion analysis can identify and solve potential and budding social problems. This study aims to realize an analysis of Chinese sentiment in social media reviews using a long short-term memory network (LSTM) model. The dataset was obtained from Sina Weibo using a web crawler and was cleaned with Pandas. First, Chinese comments regarding the legal sentencing in of Tangshan attack and Jiang Ge Case were segmented and vectorized. Then, a binary LSTM model was trained and tested. Finally, sentiment analysis results were obtained by analyzing the comments with the LSTM model. The accuracy of the proposed model has reached approximately 92%.

## 1   Introduction

Since first being introduced in China, the Internet in China has bloomed. Because of the large population in China and vast investments from the Chinese government, internet data has grown exponentially and become an important source of information. Chinese sentiment analysis bears great significance because valuable information concerning social reviews and public opinion can be obtained from these large amounts of data.

The purpose of text sentiment analysis is to determine the emotional color of text. With the rise of Web 2.0, social media platforms such as Weibo, Zhihu, and Douban have emerged and thrived as attractive online platforms for people to freely express their ideas and share their feelings. As a result, text sentiment analysis on social media content has gained extensive attention from both industrial and academic circles. As reviews on social media reflect people's emotions toward entities such as social events, products, and services, corresponding sentiment analysis can play an important role in various scenarios, such as decision making, content recommendation, and election prediction. In addition, sentiment analysis research is conducive to the development of research in other fields of NLP, such as automatic text summation [1] and opinion question answering systems [2].

In this study, an LSTM network was applied to train a binary emotion classification model for social media reviews related to criminal trials. The obtained model classifies reviews into two polarized emotions: positive and negative.

## 2   Related Work

Numerous methods have been proposed for sentiment analysis, including sentiment lexicons [3], machine learning [4] and deep learning [5]. Lexicon-based methods employ the emotional lexicon and syntactic structure of the text data to calculate emotional polarity, which requires a large number of manual labels [6]. This method greatly depends on artificial design and prior knowledge, making it difficult to popularize. Pang et al. [7] proposed the first sentiment analysis based on supervised learning methods in 2002, including naive Bayes, maximum entropy classification, and support vector machines. Subsequently, schemes based on semi-supervised methods have been developed [8, 9]. Currently, unsupervised methods, such as deep learning, are receiving increasing amounts of attention [10] owing to their flexible structure and powerful word embedding technology.

Text sentiment analysis of social media content has been extensively conducted worldwide for a vast variety of languages and topics, such as Covid-19 [11], the US presidential election [12] and the Russia-Ukraine conflict [13]. However, as far as we know, most current sentiment analysis methods are designed for English texts, and there are only limited studies focusing on Chinese sentiment analysis [14, 15], which do not consider social media reviews. Therefore, building novel LSTM models for the detection of emotion in Chinese social media reviews remains important.

# 3 Research Design and Data Collection

A flowchart of the binary emotion classification method for Chinese social media reviews is shown in Figure 1. First, review datasets are obtained from the Internet. Then, word embedding and pre-labeling are applied as text preprocessing steps to obtain word vectors and labels. Subsequently, the word vectors are used as inputs into the LSTM model to represent texts with emotional polarities. Finally, a sigmoid classifier was employed to extract the emotional information from texts and make predictions.

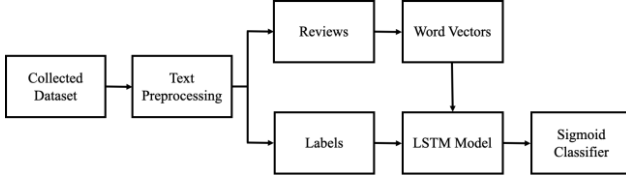

**Figure 1** Flowchart of the binary emotion classification method

## 3.1 Data Collection

We obtained the original review text regarding the sentencings of the Tangshan attack and Jiang Ge case from Sina Weibo using a web crawler. There was a corpus of 8,027 positive and 8,703 negative reviews, which was mixed with the original reviews of the Tangshan attack and used as the training and validation sets. Reviews in relation to the Jiang Ge case were used as the test set.

## 3.2 Text Preprocessing

We obtained the original review text regarding the sentencings of the Tangshan attack and Jiang Ge case from Sina Weibo using a web crawler. There was a corpus of 8,027 positive and 8,703 negative reviews, which was mixed with the original reviews of the Tangshan attack and used as the training and validation sets. Reviews in relation to the Jiang Ge case were used as the test set.

Text preprocessing is traditionally an important step of cleaning and preparing text data in NLP, including obtaining the original text, tokenization, text cleaning, and word embedding. We labeled the reviews as positive or negative using the Python class library SnowNLP. The Tangshan case dataset consisted of 7,004 positive reviews and 2,009 negative reviews, while the Jiang Ge case dataset consisted of 13,597 positive reviews and 9,633 negative reviews. Sample reviews were translated into English and are shown in Table 1. Word embedding transforms text into a form that is compatible with a neural network. In this experiment, word embedding was performed using Word2Vec. The purpose of word embedding is to represent words as high-dimensional vectors. The distance between two words in a vector space indicates their semantic or contextual similarity. Word2Vec is a packed and computationally efficient neural network in Python whose vectorized outputs are suitable input features for an LSTM model [16]. Word2Vec has two model architectures that realize the distributed representation of words: a continuous bag of words (CBOW) and a skip-gram [17]. CBOW is a three-layer neural network that predicts the current word by averaging context embeddings. Skip-gram predicts context words primarily from the center word. It is widely acknowledged that CBOW works well for small databases and trains faster, whereas Skip-Gram is more suitable for larger datasets and achieves a better performance [18].

**Table 1.** Sample positive and negative reviews from social media translated into English

| Label | Reviews |
| --- | --- |
| Positive | The sentence appears suitable; it would have been even more serious 20+ years earlier because of its negative impact on the country and even on the international community. It is right to conduct justice in a legal procedure to deter violent individuals who challenge the dignity of the law. But it also shows that we are indeed in a period of economic downturn. In the past 10+ years, such events have rarely occurred. It is understandable that people would be angry if their business is negatively impacted. The fundamental purpose of the severe punishment was to stabilize economic interests. |
| Positive | Justice is never absent and China's laws are trustworthy. |
| Negative | This matter may not be settled without being supervised by public opinion. It is supposed that the state will strengthen its efforts for full and strict governance over the Party, letting the people at the bottom believe that there are laws to follow. |
| Negative | It really makes me feel that justice has long arms and that it is only public opinion that results in strict investigation. |

## 3.3 LSTM model

LSTM [19] is a modification of a recurrent neural network (RNN) that resolves the gradient extinction problem, as shown in Figure 2. A simple RNN often fails to capture long-term correlations because the components of the gradient vector may increase or collapse exponentially over long sequences [20]. LSTM solves this problem while retaining the advantages of a simple RNN by introducing a memory unit $c_t$ and three gates: forget $f_t$, input $i_t$ and output $o_t$. These gates model the long correlation of a word sequence. Therefore, the LSTM network is well-suited for classifying, processing, and predicting time series given time lags of unknown durations. LSTM is characterized by weight matrices $W_j$, $W_j$ and biases $b_j$ for $j \in \{i, f, o, c\}$.

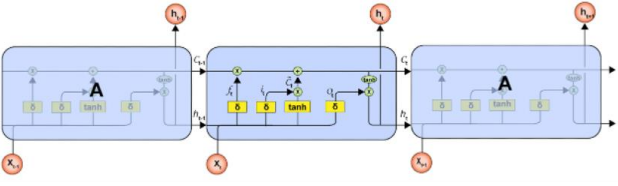

**Figure 2.** Basic structure of a simple LSTM network.

The forget gate determines which state or information from the previous cell state $c_{t-1}$ should be retained at time $t$ by calculating the remember vector $f_t$:

$$f_t = \sigma(W_f \cdot [h_{t-1}, x_t] + b_f) \quad (1)$$

where $\sigma$ is the sigmoid function; $h_{t-1}$ is the previous output, which is also the state of the hidden layer; and $x_t$ is the input at the current time step. The retained cell state after $c_{t-1}$ passes through the forget gate is $c_t^f = f_t * c_{t-1}$, where $*$ denotes the element-wise multiplication operator. The most frequently occurring state or information is retained at the forget gate, thereby improving calculation efficiency.

The input gate determines which state in $x_t$ should be retained in $c_t$:

$$i_t = \sigma(W_i \cdot [h_{t-1}, x_t] + b_i) \quad (2)$$

$$c_t = \tanh(W_c \cdot [h_{t-1}, x_t] + b_c) \quad (3)$$

where $i_t$ is input gate at the current time step and $\tanh$ is the hyperbolic tangent function. The output of the input gate is given by $c_t^i = i_t * c_t$. Therefore, the current cell state can be expressed as a combination of the outputs of the input and forget gates:

$$c_t = c_t^i + c_t^f \quad (4)$$

The output gate determines what flows from $c_{t-1}$ to the output of the current time, $h_t$:

$$o_t = \sigma(W_o \cdot [h_{t-1}, x_t] + b_o) \quad (5)$$

$$h_t = o_t * \tanh(c_t) \quad (6)$$

where $o_t$ is the output gate at the current time step and $h_t$ is the output to both the neural network and the next unit.

The LSTM builds a feedback mechanism with a memory unit and three gates. As a result, the LSTM network can capture long correlations over the entire sequence of data, which makes it a powerful tool for vast applications, such as speech recognition [21] and market prediction [22].

## 4 Experiments and Analysis

Since Chinese words are not separated by spaces, as words are in English, Chinese texts contain many ineffective stop words, which presents a great challenge in Chinese sentiment analysis [23]. In this experiment, we performed segmentation by characters rather than word segmentation by jieba because of the limited size of the data. Then, we applied word embedding with a dimension of 300. We trained the LSTM model using the activation function. The dropout technique and early stopping function were applied to prevent overfitting during training. We used Adam, an optimization algorithm based on a step objective function, as the optimizer. The accuracy and training loss during the training are shown in Figure 3.

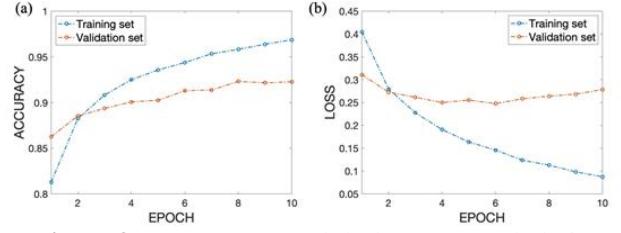

**Figure 3.** (a) Accuracy and (b) loss per epoch during training.

Results from testing the model on the reviews of the Jiang Ge trial are outlined in Table 2. Satisfactory accuracy was obtained, which demonstrates that the LSTM model was able to classify the emotions of the reviews accurately.

**Table 2.** Evaluation indexes of the proposed LSTM model.

| Loss | MAE | Accuracy | Precision | Recall |
|------|-----|----------|-----------|--------|
| 0.2627 | 0.0916 | 92.06% | 92.08% | 92.04% |

## 5 Conclusions

In this study, we introduced the principle of LSTM and presented a binary Chinese sentiment analysis model based on LSTM, with a focus on social media reviews of popular criminal trials. The proposed model used Adam as the optimization method. Syntactic and semantic information were extracted using the word embedding technique Word2Vec. The experimental results demonstrate excellent performance and suggest that LSTM networks are promising for sentiment analysis of social media reviews. This model may be applicable to public opinion mining and supervision, which is crucial for the rule of law in a country.

*Big Data Analytics and Computational Intelligence (ICBDAC)*. 2017. IEEE.